\documentclass[runningheads]{llncs}
\usepackage{graphicx}
\usepackage{fancyvrb}

\title{A tale of two toolkits, report the first: benchmarking time series classification algorithms for correctness and efficiency}
\author{Anthony Bagnall$^{1}$ \and Franz J.~Kir\'{a}ly$^{2}$  \and  Markus L{\"o}ning$^{2}$ \and Matthew Middlehurst$^{1}$ \and George Oastler$^{1}$}
\authorrunning{A. Bagnall {\em et al.}}
\titlerunning{Time series classification in sktime}

\institute{University of East Anglia
\and
University College London \& The Alan Turing Institute}

\begin{document}

\maketitle
\begin{abstract}
sktime is an open source, Python based, sklearn compatible toolkit for time series analysis developed by researchers at the University of East Anglia (UEA), University College London and the Alan Turing Institute. A key initial goal for sktime was to provide time series classification functionality equivalent to that available in a related java package, tsml, also developed at UEA. We describe the implementation of six such classifiers in sktime and compare them to their tsml equivalents. We demonstrate correctness through equivalence of accuracy on a range of standard test problems and compare the build time of the different implementations. We find that there is significant difference in accuracy on only one of the six algorithms we look at (Proximity Forest~\cite{lucas19proximity}). This difference is causing us some pain in debugging. We found a much wider range of difference in efficiency. Again, this was not unexpected, but it does highlight ways both toolkits could be improved.

\end{abstract}

\section{Introduction}
sktime\footnote{https://github.com//alan-turing-institute/sktime} is an open source, Python based, sklearn compatible toolkit for time series analysis developed by researchers at the University of East Anglia, University College London and the Alan Turing Institute. sktime is designed to provide a unifying API for a range of time series tasks such as annotation, prediction and forecasting (see~\cite{loning19sktime} for a description of the overarching design of sktime). A key prediction component is algorithms for time series classification (TSC). We have implemented a broad range of TSC algorithms within sktime. This technical report provides evidence of the correctness of the implementation and assesses the time efficiency of these classifiers by comparing accuracy and run time to the Java versions within the tsml toolkit\footnote{https://github.com/uea-machine-learning/tsml} used in an extensive experimental comparison published in 2017~\cite{bagnall17bakeoff}. This is a work in progress, and this document will be updated as we improve and add functionality.

\section{sktime structure}

sktime is organised into high level packages to reflect the different tasks, shown in Figure~\ref{fig:packages}. We are concerned with the packages \texttt{classifiers}, the components of which make use of \texttt{distance\_measures} and \texttt{transformers}. Note that in the \texttt{dev} branch there is a package \texttt{contrib}. This contains implementations that have not been fully integrated yet and bespoke code to run experiments.
\begin{figure}[ht]
	\centering
        \includegraphics[height =8cm]{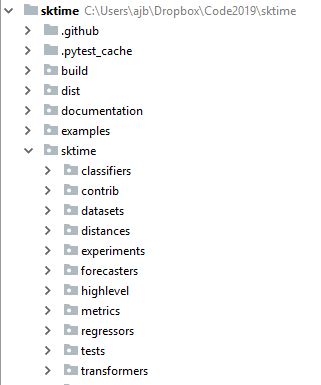}
    \caption{Overview of the current sktime file structure (sktime version 0.3).}
    \label{fig:packages}
\end{figure}
Classifiers are grouped based on their core transformation technique.
\begin{figure}[ht]
	\centering
        \includegraphics[height =8cm]{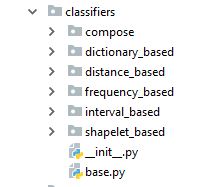}
    \caption{Overview of the current sktime classifier packages (sktime version 0.3).}
    \label{fig:packages}
    \end{figure}
The taxonomy of classifiers is taken from the comparative study presented in~\cite{bagnall17bakeoff}. Currently, the following classifiers are available.

\texttt{interval\_based} is a package for classifiers built on intervals of the series. Currently, it contains a single classifier the Time Series Forest~\cite{deng13forest}, in file \texttt{tsf.py}. This is evaluated in Section~\ref{sec:interval}. Spectral based classifiers are located in the package \texttt{frequency\_based}. The  random interval spectral ensemble (RISE)~\cite{lines18hive} is as yet the only concrete implementation of this type and it can be found in \texttt{rise.py} (see Section~\ref{sec:frequency}).  \texttt{dictionary\_based} contains classifiers that form histograms of discretised words. We have the Bag of SFA Symbols (BOSS)~\cite{schafer15boss} classifier, implemented as \texttt{BOSSEnsemble} in \texttt{boss.py}, with experiments described in Section~\ref{sec:dictionary}. Shapelets are discriminatory subseries, and classifiers that use them are in the \texttt{shapelet\_based} package. We evaluate a version of the shapelet transform classifier~\cite{bostrom17binary} implemented in the file \texttt{stc.py}.

\noindent In \texttt{time\_series\_neigbours.py} there is \texttt{KNeighborsTimeSeriesClassifier}, an sklearn based nearest neighbour (NN) classifier that can be configured to use a range of distance measures, including: dynamic time warping (DTW); derivative  dynamic time warping (DDTW)~\cite{gorecki13derivative}; weighted DTW~\cite{jeong11weighted}; longest common subsequence; edit distance with penalty (ERP); and move-split-merge~\cite{stefan13move-split-merge}. Until recently, DTW with 1-NN was by far the most commonly used benchmark algorithm for time series classification. These distance functions are available in the package \texttt{distances}.
we evaluate the  \texttt{ElasticEnsemble}~\cite{lines15elastic}, and ensemble of nearest neighbour classifiers using elastic distance measures, which is in \texttt{elastic\_ensemble.py}. Furthermore, in \texttt{proximity.py} there is the \\ \texttt{ProximityForest}~\cite{lucas19proximity}, a tree based classifier using a range of distance measures. DTW, EE and PF classifiers are evaluated in Section~\ref{sec:distance}.  The sktime experiments described in the following sections can be reproduced using the code in \texttt{contrib/basic\_benchmarking.py}.

\section{Interval Based: Time Series Forest (TSF)~\cite{deng13forest}}
\label{sec:interval}
TSF\footnote{http://www.timeseriesclassification.com/algorithmdescription.php?algorithm\_id=8} is probably the simplest bespoke classifier and hence was the first algorithm we implemented. TSF is an ensemble of tree classifiers, each constructed on a different feature space. For any given ensemble member, a set of random intervals is selected, and three summary statistics (mean, standard deviation and slope) are calculated for each interval. The statistics for each interval are concatenated to form the new feature space.

In sktime, there are two ways of constructing a TSF estimator. The hard coded implementation \texttt{TimeSeriesForest} is in the file
\texttt{tsf.py}. An example usage, including data loading and building, assuming \texttt{full\_path} is where the problems reside\footnote{ https://github.com/alan-turing-institute/sktime/blob/master/examples/loading\_data.ipynb for  examples of how to load data}, is as follows:

\begin{Verbatim}[frame=single]
from sktime.utils.load_data import load_from_tsfile_to_dataframe
                                                    as load_ts
import sktime.classifiers.interval_based.tsf as ib
#method 1
tsf=ib.TimeSeriesForest(n_trees=100)
trainX, trainY = load_ts(full_path + '_TRAIN.ts')
testX, testY = load_ts(full_path  + '_TEST.ts')
tsf.fit(trainX, trainY)
predictions=tsf.predict(testX)
probabilities=tsf.predict_proba(testX)
\end{Verbatim}
Alternatively, you can configure the \texttt{TimeSeriesForestClassifier} in \\ \texttt{classifiers.compose.ensemble.py} as TSF as follows
\begin{Verbatim}[frame=single]
#method 2
steps = [
 ('segment', RandomIntervalSegmenter(n_intervals='sqrt')),
 ('transform', FeatureUnion([
    ('mean', RowwiseTransformer(FunctionTransformer(func=np.mean,
                                validate=False))),
    ('std', RowwiseTransformer(FunctionTransformer(func=np.std,
                                validate=False))),
    ('slope', RowwiseTransformer(
                FunctionTransformer(func=time_series_slope),
                                    validate=False))
 ])),
 ('clf', DecisionTreeClassifier())
]
base_estimator = Pipeline(steps)
tsf = TimeSeriesForestClassifier(base_estimator=base_estimator,
                                 n_estimators=100)
\end{Verbatim}
The \texttt{validate=False} is to stop the built in methods using the sktime data format incorrectly (and to suppress annoying warnings). It will be set by default to \texttt{False} in the next release. The configurable version allows for the easy formulation of TSF like variants and other transformation based ensembles.  However, this comes at the cost of efficiency. To verify the implementations, we compare the tsml java version\footnote{https://github.com/uea-machine-learning/tsml/} used in the bake off~\cite{bagnall17bakeoff} to the two sktime versions.

We aim to show there is no difference in accuracy between the three implementations. We evaluate these three classifiers on 30 stratified resamples on 111 of the UCR archive~\cite{dau18archive}.

Full results are available on the associated website\footnote{http://www.timeseriesclassification.com/sktime.php}. Figure~\ref{fig:tsf_cd} shows the critical difference diagrams~\cite{demsar06comparisons} for the three versions for accuracy, balanced accuracy, area under the receiver operator curve (AUC) and the negative log likelihood (NLL). Solid bars indicate 'cliques', within which there is no significant difference between members. A fuller explanation is provided in~\cite{bagnall18rotf}. Figure~\ref{fig:tsf_cd} demonstrates there is no significant difference in between the classifiers in accuracy, balanced accuracy and NLL. The only observed difference is that sktime1 is significantly better than sktime2 with AUC. This may be explainable by the voting method employed by the two algorithms being different, although we have not confirmed it. We suspect this is just a chance result. The actual differences in AUC are very small.

\begin{figure}[!ht]
	\centering
\begin{tabular}{cc}
       \includegraphics[width =6.5cm,trim={4cm 10cm 2cm  3cm},clip]{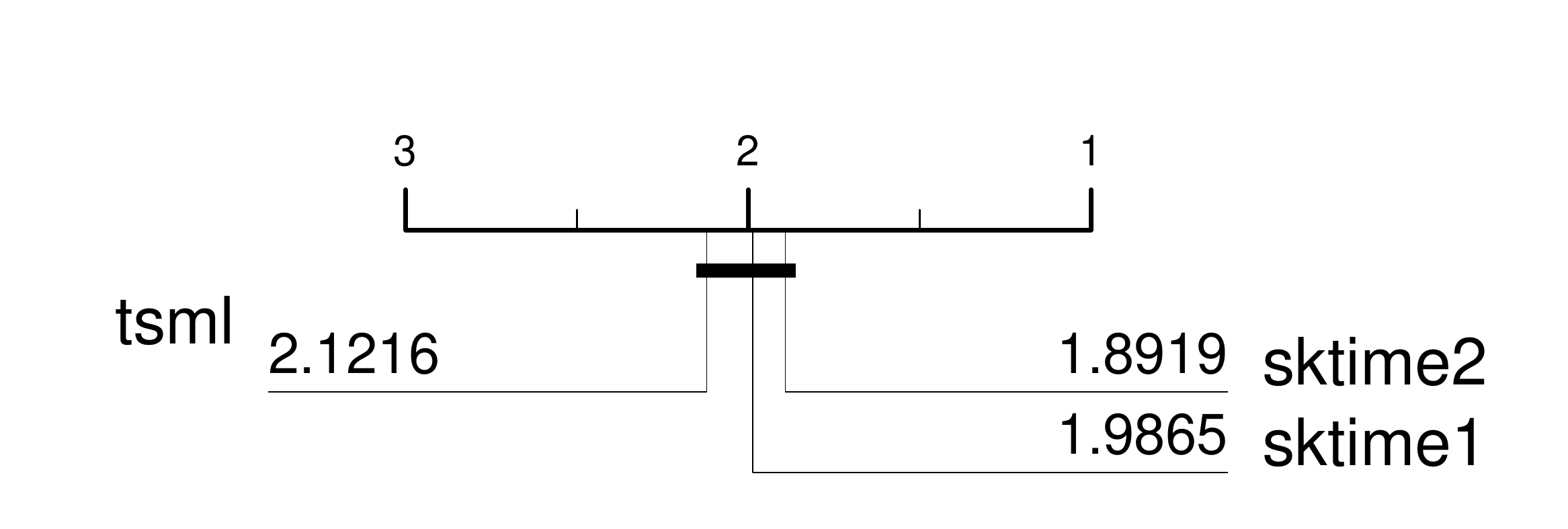}              	
&
       \includegraphics[width =6.5cm, trim={4cm 10cm 2cm  3cm},clip]{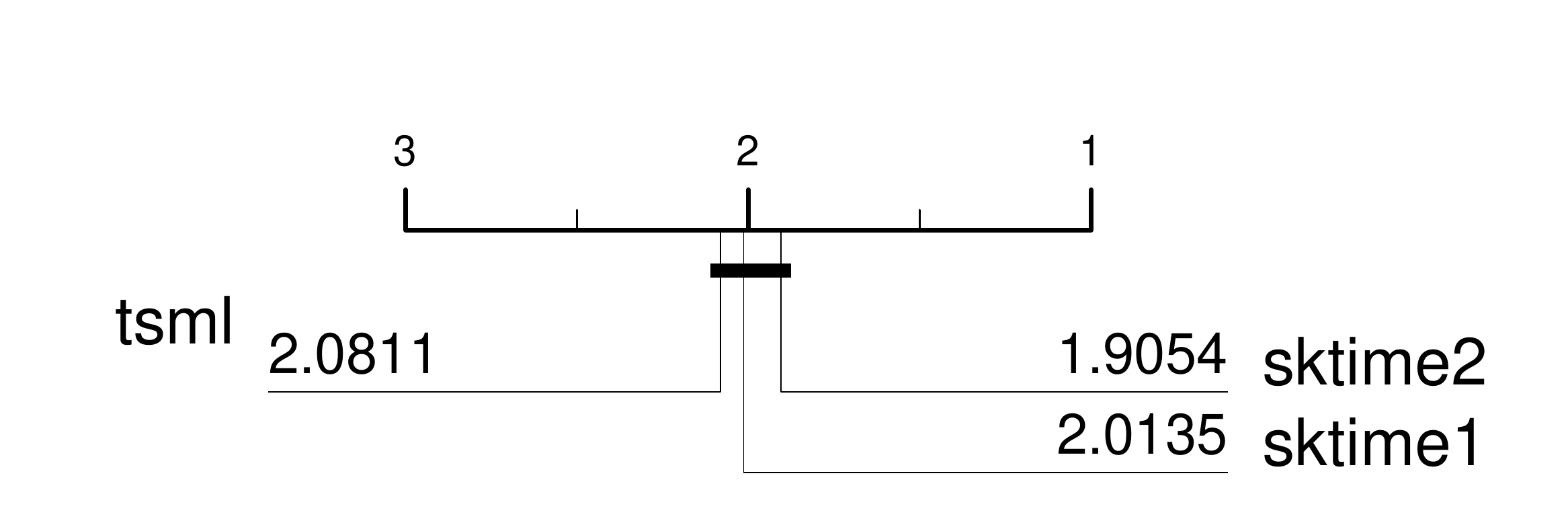}  \\
(a) Accuracy & (b) Balanced Accuracy \\
       \includegraphics[width =6.5cm, trim={4cm 10cm 2cm  3cm},clip]{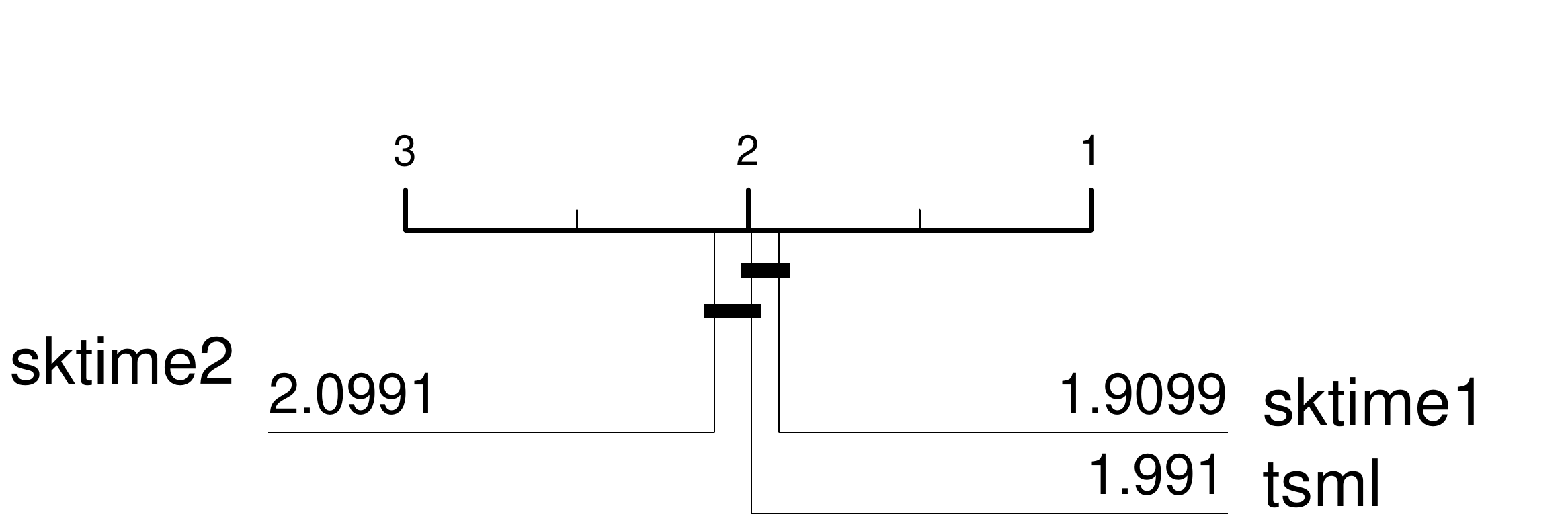}
&
       \includegraphics[width =6.5cm, trim={4cm 10cm 2cm  3cm},clip]{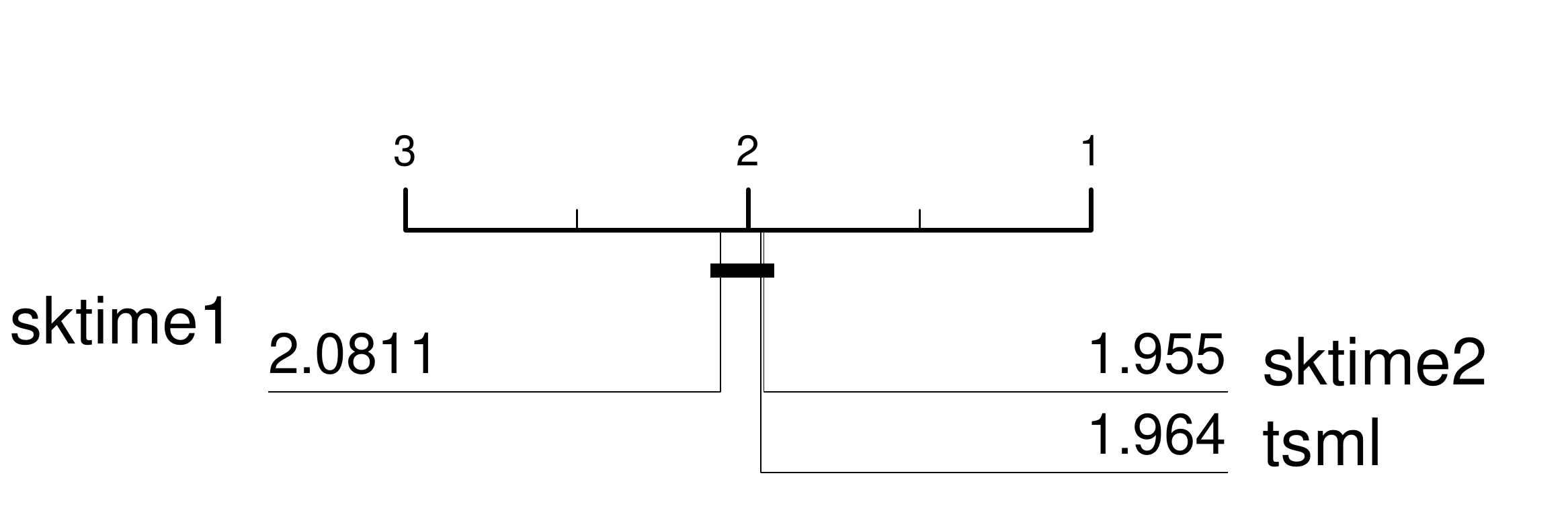}              	\\
(c) AUC & (d) NLL \\          
       \end{tabular}
       \caption{Critical difference diagrams for three TSF implementations. Two from sktime (sktime1 using method 1 and sktime2 using method 2) and one from tsml. There is no significant difference between the three classifiers (as tested with a pairwise wilcoxon test with Holm correction, $\alpha=0.05$).}
       \label{fig:tuned}
\end{figure}

\begin{figure}[!htb]
\centering
    \includegraphics[width=\textwidth, trim={0cm 0cm 0cm  3cm},clip]{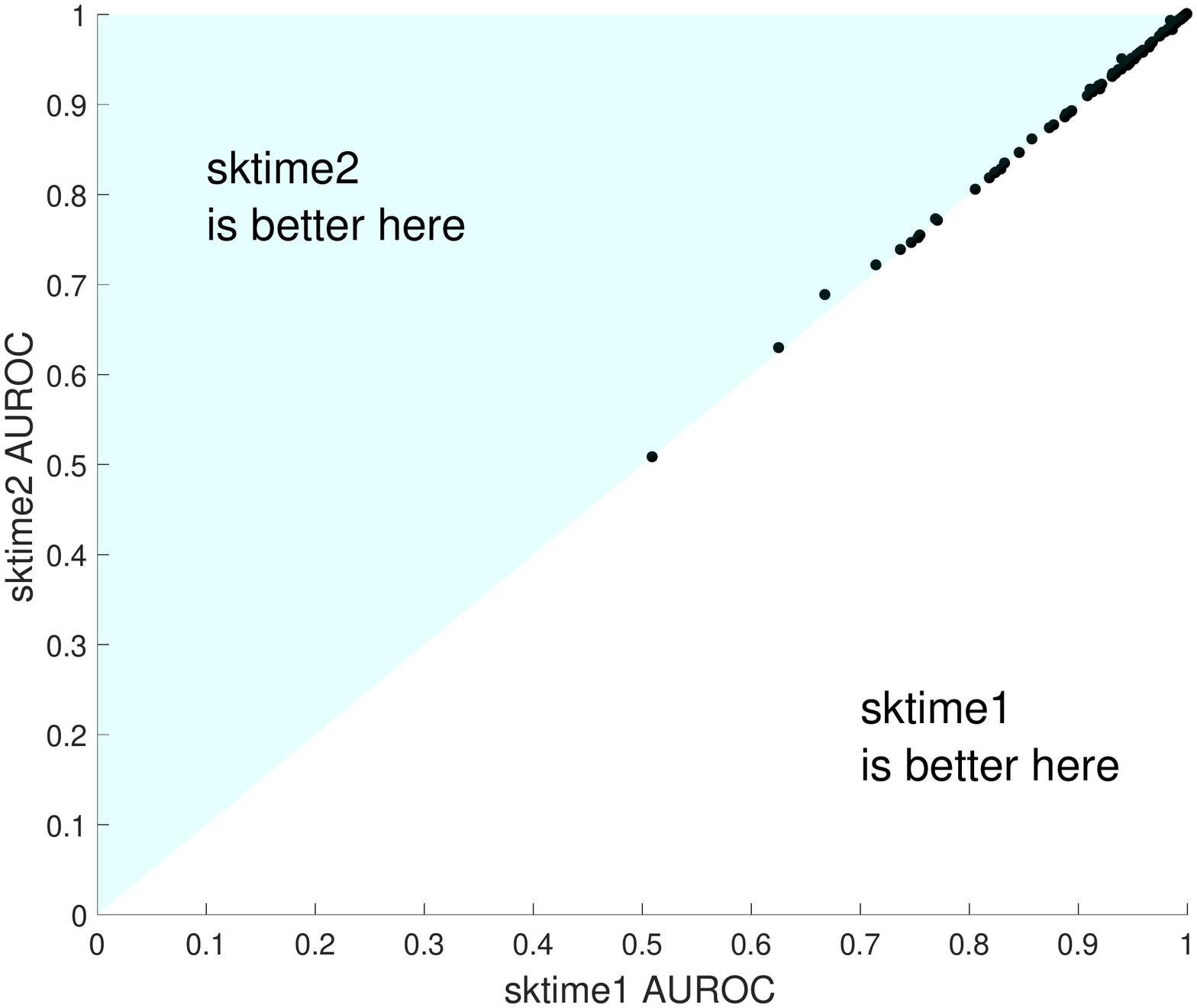}
    \caption{TSF sktime1 vs sktime2 scatterplot of AUC. sktime2 Wins/Draw/Losses: 41/5/65}
    \label{fig:tsf_pairwise1}
 \end{figure}

\begin{figure}[!htb]
\centering
\includegraphics[width=\textwidth, trim={0cm 0cm 0cm  3cm},clip]{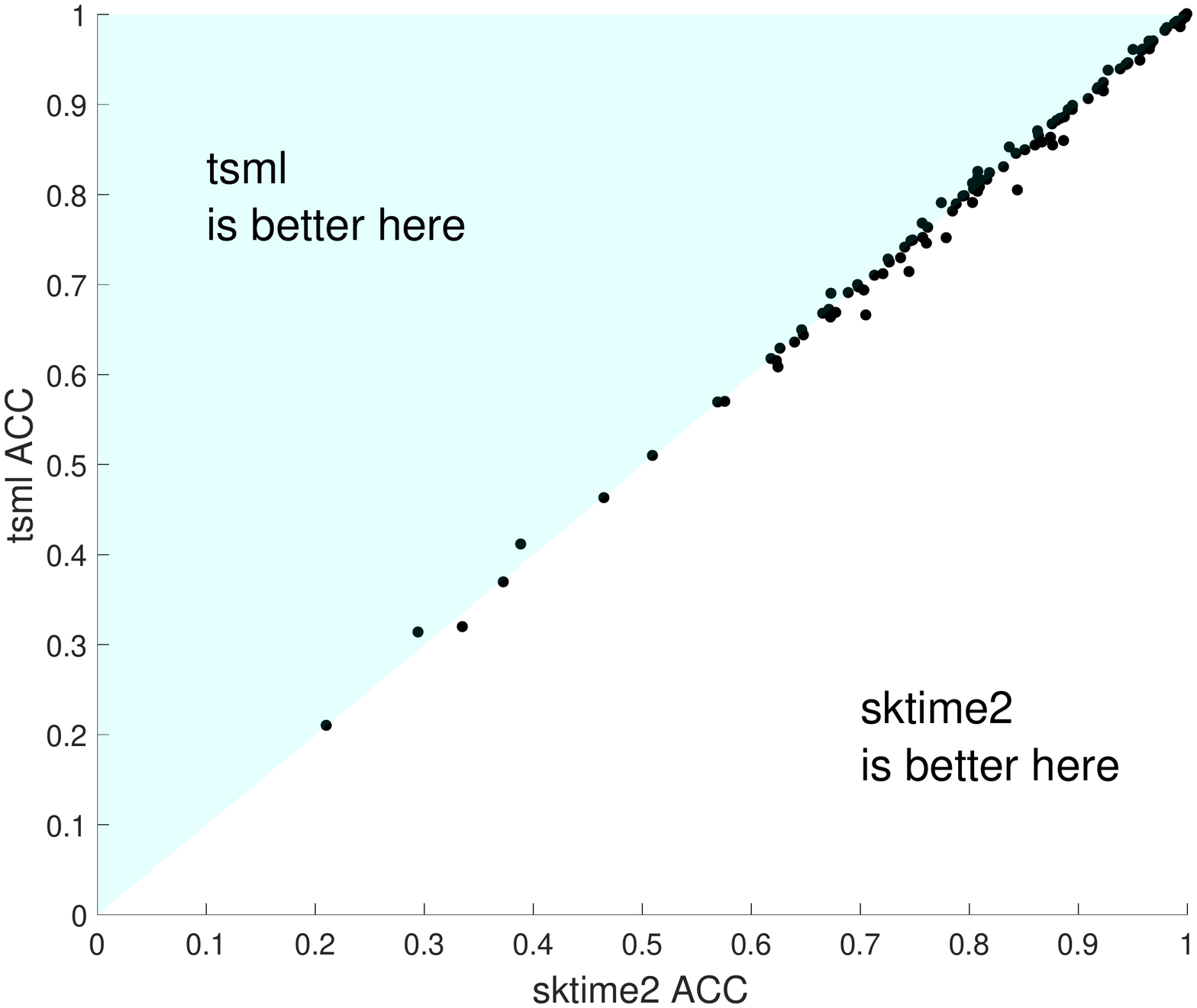}
    \caption{TSF tsml vs sktime2 scatterplot of test accuracies. sktime2 Wins/Draw/Losses: 60/3/48.}
    \label{fig:tsf_pairwise2}
\end{figure}

\begin{figure}[!h]
	\centering
        \includegraphics[width=\textwidth, trim={0cm 0cm 0cm  0cm},clip]{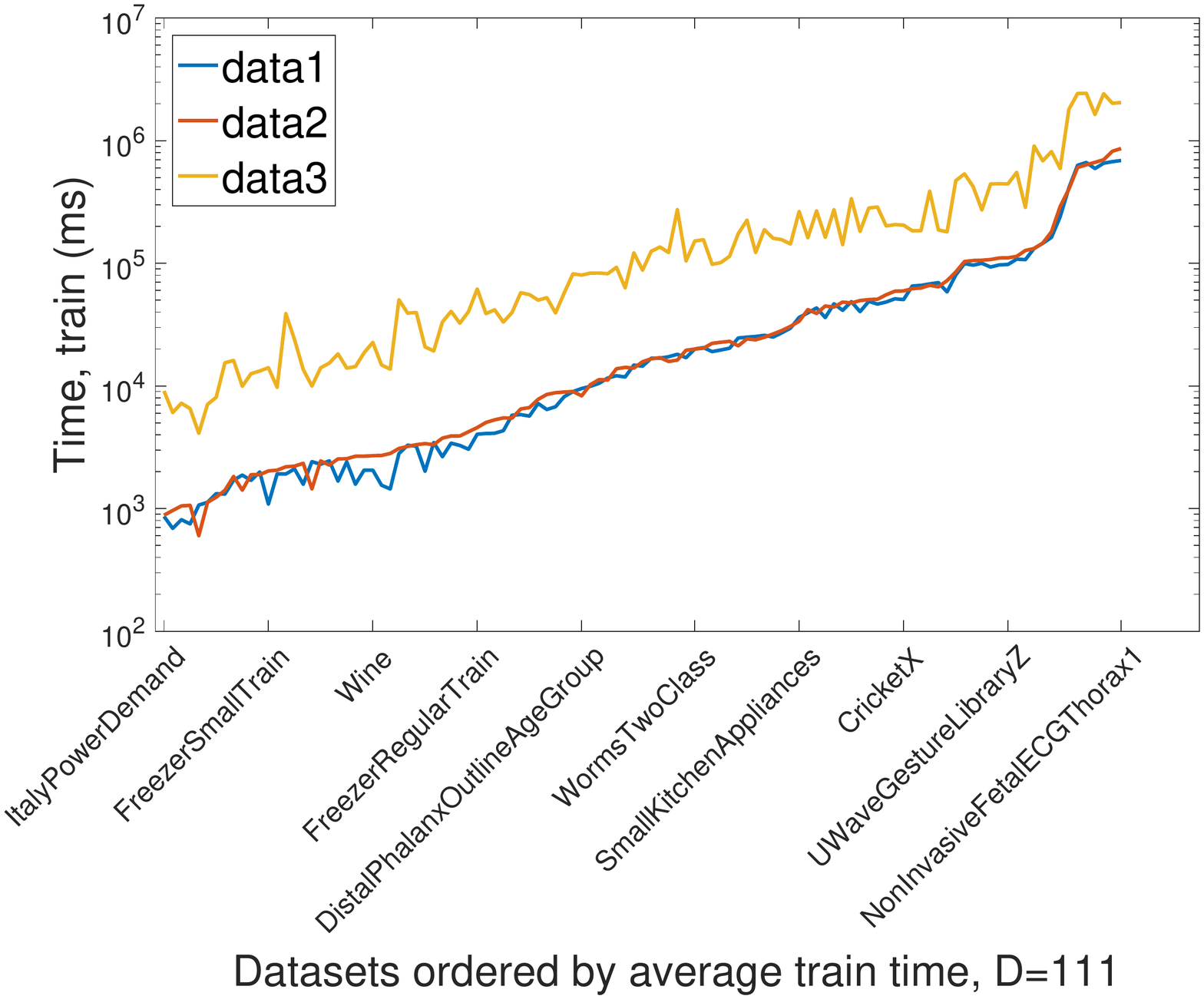}
    \caption{Train time for three TSF classifiers: sktime1 (data1); tsml (data2); and sktime2 (data3).}
    \label{fig:tsf_timing}
\end{figure}

Figures~\ref{fig:tsf_pairwise1} and~\ref{fig:tsf_pairwise2} shows the pairwise scatter plots for sktime2 vs tsml and sktime1.
\begin{figure}[!htb]
\centering
    \includegraphics[width=\textwidth, trim={0cm 0cm 0cm  3cm},clip]{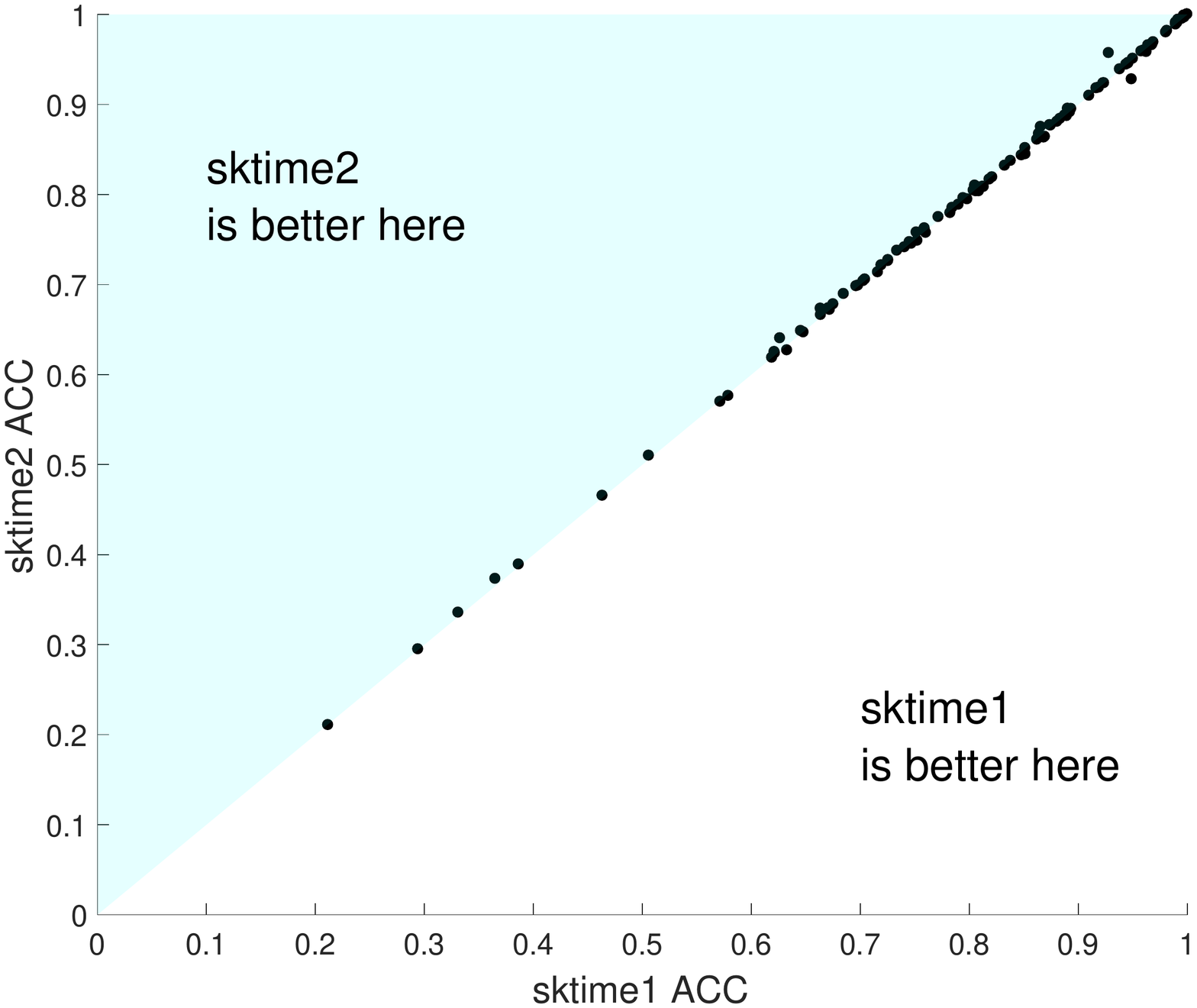}
    \caption{TSF sktime1 vs sktime2 scatterplot of test accuracies. sktime2 Wins/Draw/Losses: 59/5/47.}
    \label{fig:tsf_pairwise1}
 \end{figure}
Timing comparison for the three classifiers is shown in Figure~\ref{fig:tsf_timing}.

The configurable version is an order of magnitude slower than the fixed version, which is approximately equivalent to the java based tsml. We are not entirely sure as to the reason for this slowdown with the composite. We suspect it is due to excessive unpacking and packing of data into pandas, but it merits further investigation and detailed profiling, particularly as the accuracy is marginally, although not significantly, higher.

\section{Frequency Based: RISE~\cite{lines18hive}}
\label{sec:frequency}
RISE is similar to TSF in that it is an ensemble of decision tree classifiers. The difference lies in the feature space used for each base classifier. RISE selects a single random interval for each base classifier, then transforms the time series in that interval into the power spectrum and autocorrelation features. As with TSF, there are two ways of building RISE.

\newpage
\begin{Verbatim}[frame=single]
import sktime.classifiers.interval_based.tsf as ib
from sklearn.preprocessing import FunctionTransformer
from sklearn.tree import DecisionTreeClassifier
from statsmodels.tsa.stattools import acf
from sktime.transformers.compose import RowwiseTransformer
from sktime.transformers.segment import RandomIntervalSegmenter
from sktime.transformers.compose import ColumnTransformer
from sktime.transformers.compose import Tabulariser
from sktime.pipeline import Pipeline
from sktime.pipeline import FeatureUnion
from sktime.classifiers.compose import TimeSeriesForestClassifier

#method 1: fixed classifier
rise = fb.RandomIntervalSpectralForest(n_trees=100)

\end{Verbatim}

\begin{figure}[!htb]
	\centering
        \includegraphics[width=\textwidth, trim={0cm 10cm 0cm  2cm},clip]{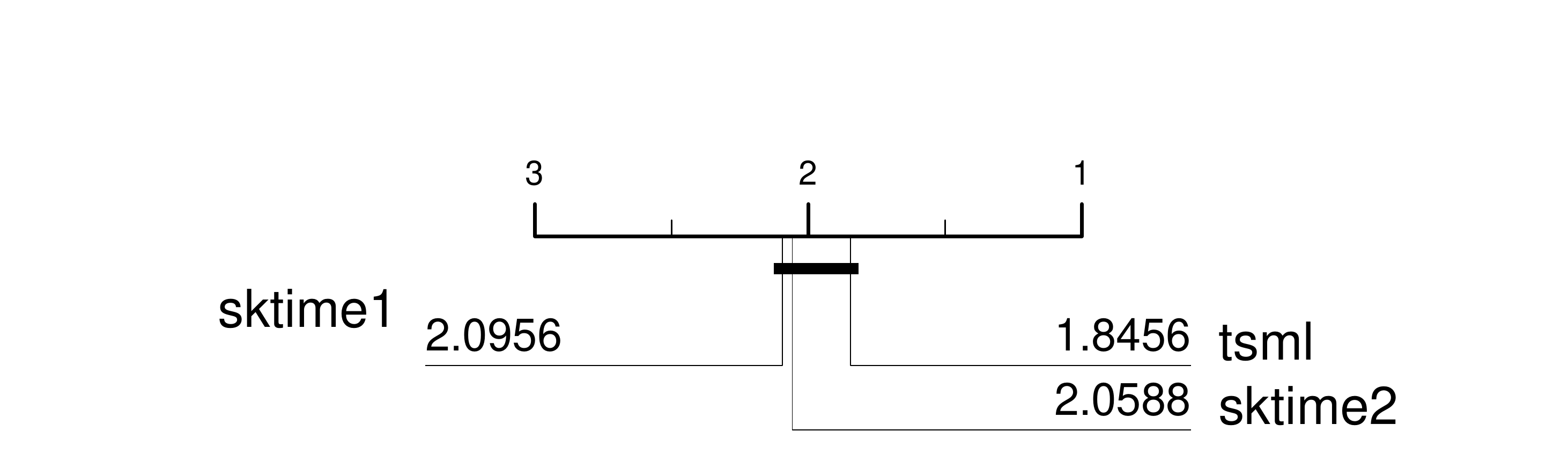}
    \caption{Average ranks for three RISE implementations. Two from sktime (sktime1 using method 1 and sktime2 using method 2) and one from tsml. There is no significant difference between the three classifiers (as tested with a pairwise wilcoxon test with Holm correction, $\alpha=0.05$).}
    \label{fig:rise_cd}
\end{figure}

\begin{figure}[!htb]
\centering
    \includegraphics[width=\textwidth, trim={0cm 0cm 0cm  3cm},clip]{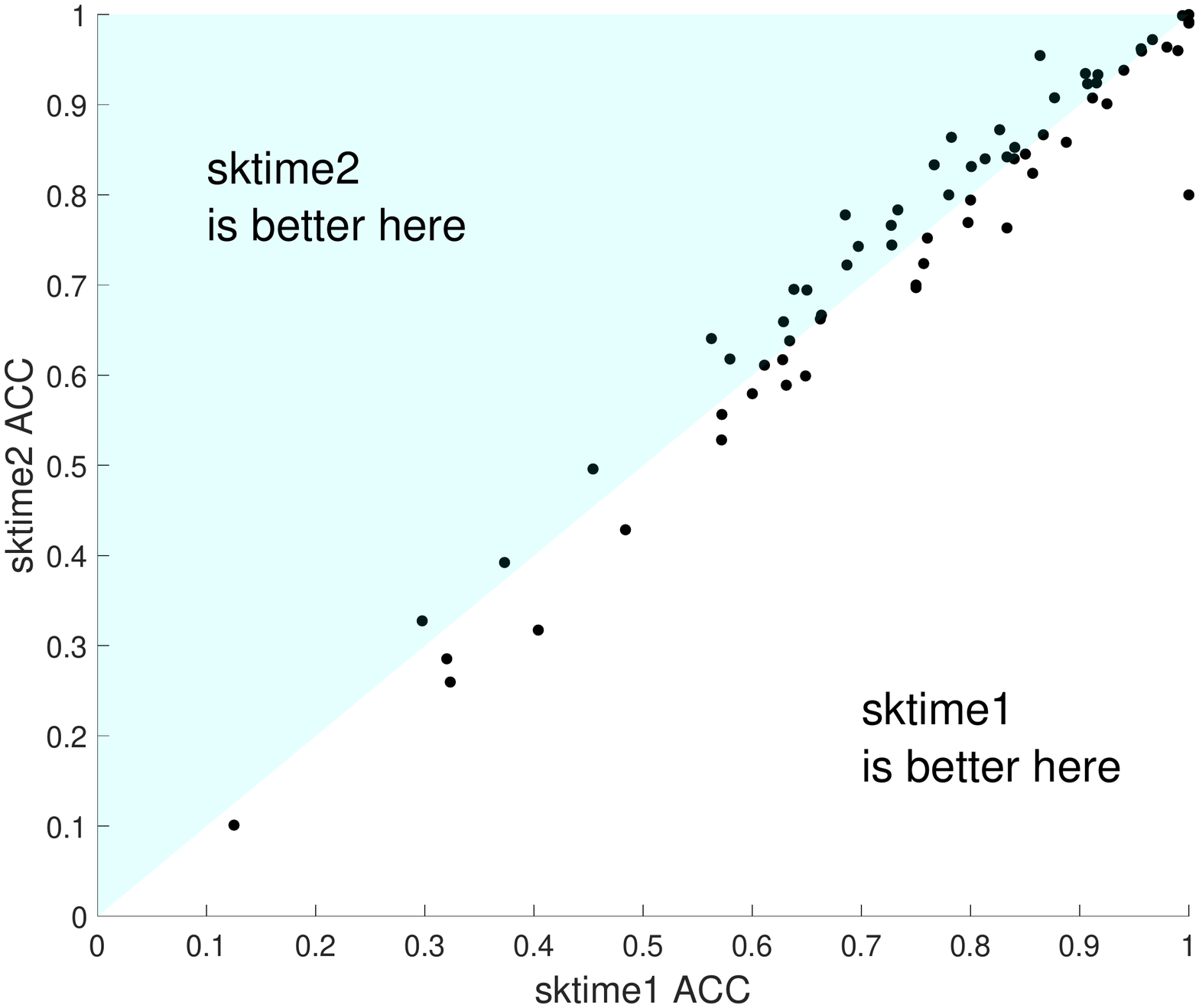}
    \caption{RISE sktime2 vs sktime1 scatterplot of test accuracies. sktime2 Wins/Draw/Losses: 34/5/29}
    \label{fig:rise_pairwise1}
 \end{figure}

\begin{figure}[!htb]
\centering
\includegraphics[width=\textwidth, trim={0cm 0cm 0cm  3cm},clip]{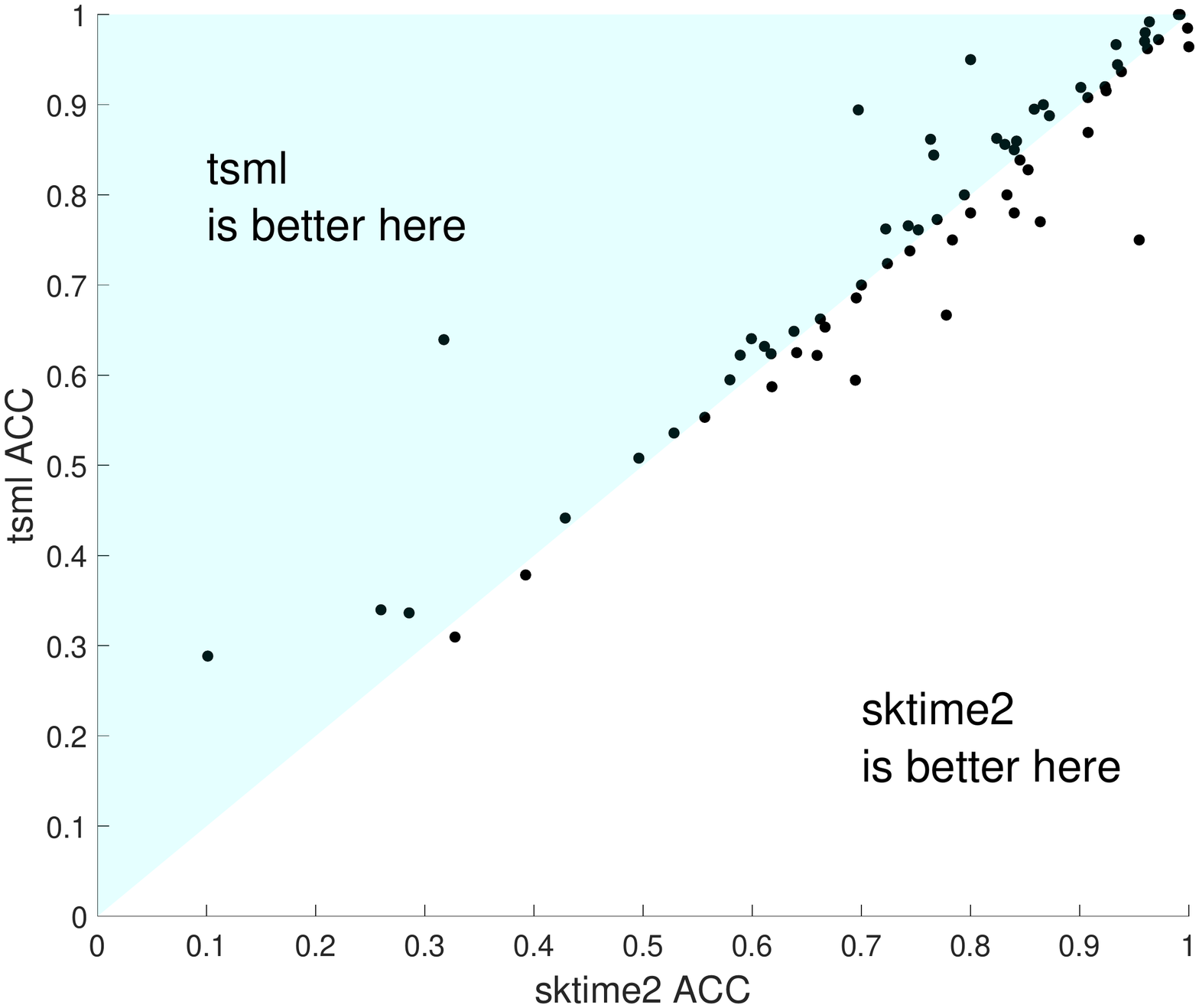}
    \caption{RISE tsml vs sktime2 scatterplot of test accuracies. sktime2 Wins/Draw/Losses: 25/5/38}
    \label{fig:rise_pairwise2}
\end{figure}

\begin{Verbatim}[frame=single]
#method 2: configurable classifier
steps = [
 ('segment',RandomIntervalSegmenter(n_intervals=1, min_length=5)),
 ('transform',FeatureUnion([
  ('acf',RowwiseTransformer(FunctionTransformer(func=acf_coefs,
                                               validate=False))),
  ('ps',RowwiseTransformer(FunctionTransformer(func=powerspectrum,
                                               validate=False)))
 ])),
  ('tabularise', Tabulariser()),
  ('clf', DecisionTreeClassifier())
  ]
base_estimator = Pipeline(steps)
rise = TimeSeriesForestClassifier(base_estimator=base_estimator,
                                    n_estimators=100)

def acf_coefs(x, maxlag=100):
    x = np.asarray(x).ravel()
    nlags = np.minimum(len(x) - 1, maxlag)
    return acf(x, nlags=nlags).ravel()

def powerspectrum(x, **kwargs):
    x = np.asarray(x).ravel()
    fft = np.fft.fft(x)
    ps = fft.real * fft.real + fft.imag * fft.imag
    return ps[:ps.shape[0] // 2].ravel()
\end{Verbatim}
The composite version requires the definition of the acf and power spectrum functions. We will make this better encapsulated in the next release. To validate these implementations, we again benchmark against the tsml Java version, using RISE with 50 trees. Using the built in numpy functions for sktime2 creates a problem with some datasets: if a zero variance interval is passed to the numpy functions, an exception is thrown. In sktime1 and tsml we can manually adjust for this circumstance. The nature of the composite sktime2 makes this harder to manage. This means we are only able to compare on 68 datasets. Figures~\ref{fig:rise_cd},~\ref{fig:rise_pairwise1} and~\ref{fig:rise_pairwise2} show there is no significant difference in accuracy between the three classifiers. However, there is greater variance in the difference in accuracy than observed for TSF.

\begin{figure}[!htb]
	\centering
        \includegraphics[width=\textwidth, trim={0cm 0cm 0cm  0cm},clip]{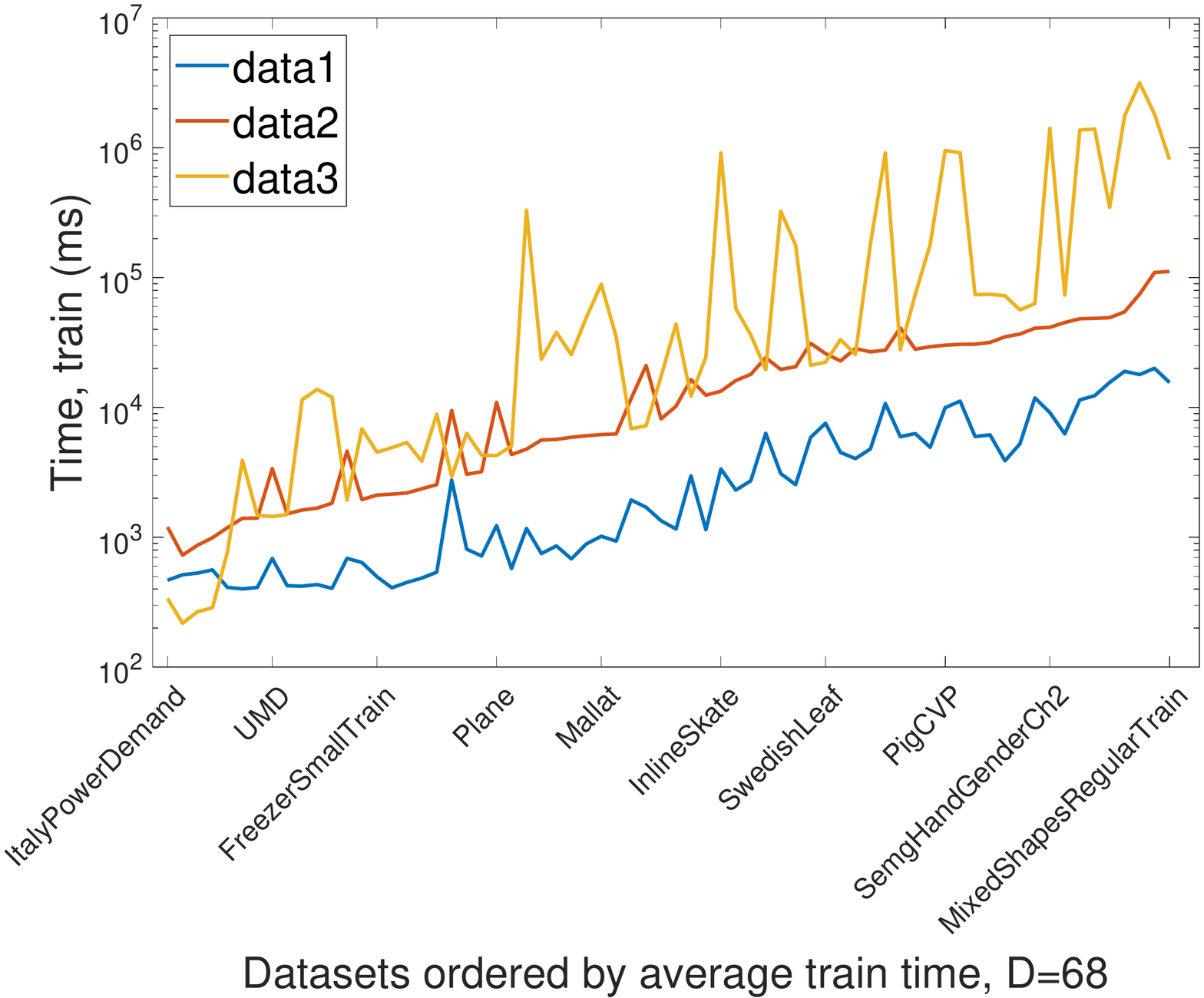}
    \caption{Train time for three RISE classifiers: sktime1 (data2); tsml (data3); and sktime2 (data1).}
    \label{fig:rise_timing}
\end{figure}

The timing comparison for RISE, shown in Figure~\ref{fig:rise_timing}, highlights two curious characteristics. Firstly, sktime2 is an order of magnitude faster than both sktime1 and tsml. We believe this is due to to the efficient numpy implementations used, although it merits further investigation. Secondly, tsml has much higher variance than the sktime implementations, and seems to take an unexpectedly long time on some small problems. This also merits further investigation.

\section{Dictionary Based: BOSS~\cite{schafer15boss}}
\label{sec:dictionary}

The BOSS implementation contains two components, \texttt{BOSSIndividual} and the \texttt{BOSSEnsemble} (what we refer to as BOSS).
Individual BOSS classifiers create a histogram of words from each series using the \texttt{SFA} transform over sliding windows.
A one nearest neighbour classifier using a bespoke BOSS distance is then used for classification.
The \texttt{BOSSEnsemble} is an ensemble of \texttt{BOSSIndividual} classifiers.
The ensemble members are selected through a grid search of individual BOSS parameters, with classifiers below 92\% accuracy of the best classifier removed.

We also implement a faster and more configurable version of the classifier cBOSS~\cite{middlehurst2019scalable}, replacing the grid search with a filtered random selection.
This alternative method can be employed with the \texttt{BOSSEnsemble} by setting the \texttt{randomised\_ensemble} parameter to true.
Two parameters are associated with this version: the number of individual BOSS classifiers to be built \texttt{n\_parameter\_samples} and the maximum number of classifiers in the ensemble \texttt{max\_ensemble\_size}.
cBOSS is also contractable, allowing a unit of time in minutes using the \textit{time\_limit} parameter to replace the \textit{n\_parameter\_samples}.

Examples of how to construct BOSS and cBOSS are demonstrated in the following code sample.
\newpage
\begin{Verbatim}[frame=single]
import sktime.classifiers.dictionary_based.boss as db

# By default the original BOSS algorithm is setup
boss = db.BOSSEnsemble()

# Configuration for recommended cBOSS settings
c_boss = db.BOSSEnsemble(randomised_ensemble=True,
                   n_parameter_samples=250, max_ensemble_size=50)

# cBOSS contracted for 1 hour, input time must be in minutes
c_boss_contract = db.BOSSEnsemble(randomised_ensemble=True,
                             time_limit=60, max_ensemble_size=50)
\end{Verbatim}

Our primary goal is to test the correctness and efficiency of the standard implementations. To this end, we compare the implementations of BOSS in sktime and tsml. Figure~\ref{fig:boss_pairwise} plots the accuracy of the two classifiers on 92 datasets. The accuracies are very similar (63 are identical and the differences are insignificant. We expected less variation in accuracy with the nearest neighbour classifier than the tree based classifiers, because there are fewer stochastic elements. This is indeed the case. However, the timing results displayed in Figure~\ref{fig:boss_timing} show how much slower the sktime implementation is: approximately 10-40 times slower than tsml. This was also not unexpected. The implementation is mapped directly from Java and does not exploit any of the efficiency improvements that can be used in python. This is required future work.
\begin{figure}[!h]
\centering
    \includegraphics[width=\textwidth, trim={0cm 0cm 0cm  3cm},clip]{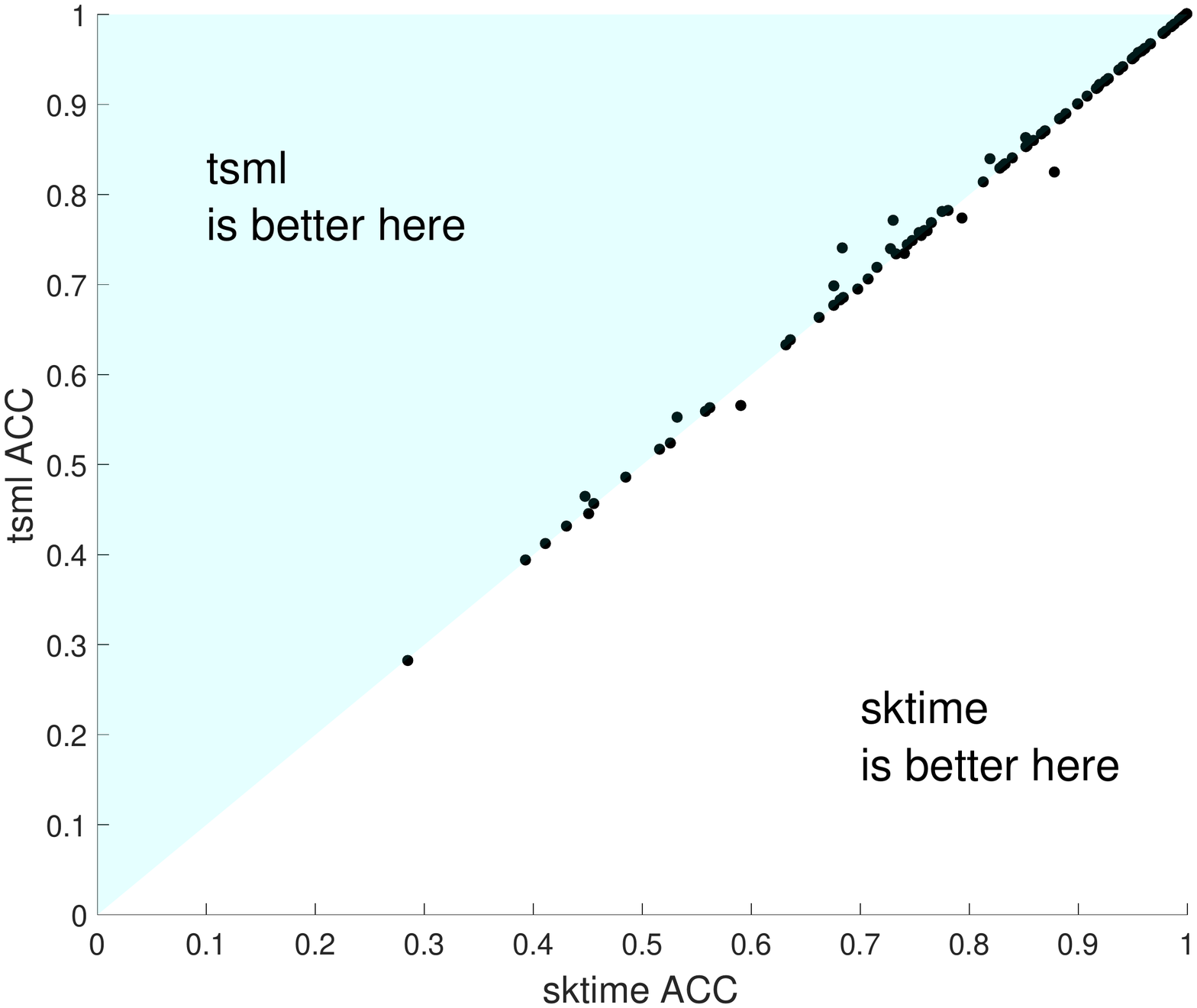}
    \caption{BOSS sktime vs tsml scatterplot of test accuracies. sktime Wins/Draw/Losses: 12/63/12.}
    \label{fig:boss_pairwise}
 \end{figure}
\begin{figure}[!h]
	\centering
        \includegraphics[width=\textwidth, trim={0cm 0cm 0cm  0cm},clip]{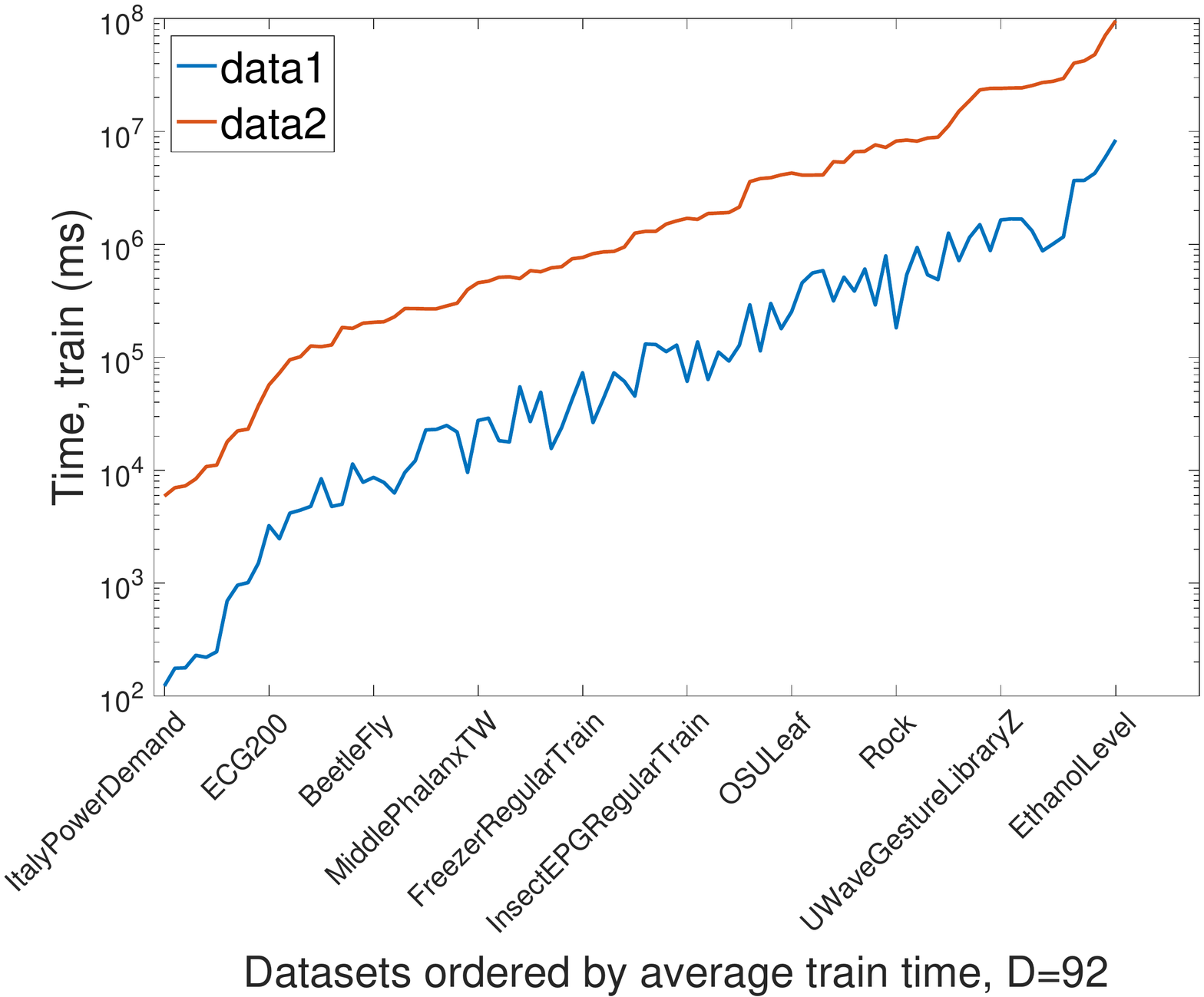}
    \caption{Train time for two BOSS classifiers: sktime (data2) and tsml (data1).}
    \label{fig:boss_timing}
\end{figure}
\section{Shapelet Based: Shapelet Transform Classifier (STC)~\cite{hills14shapelet}}
\label{sec:shapelets}
\begin{figure}[!h]
\centering
    \includegraphics[width=\textwidth, trim={0cm 0cm 0cm  3cm},clip]{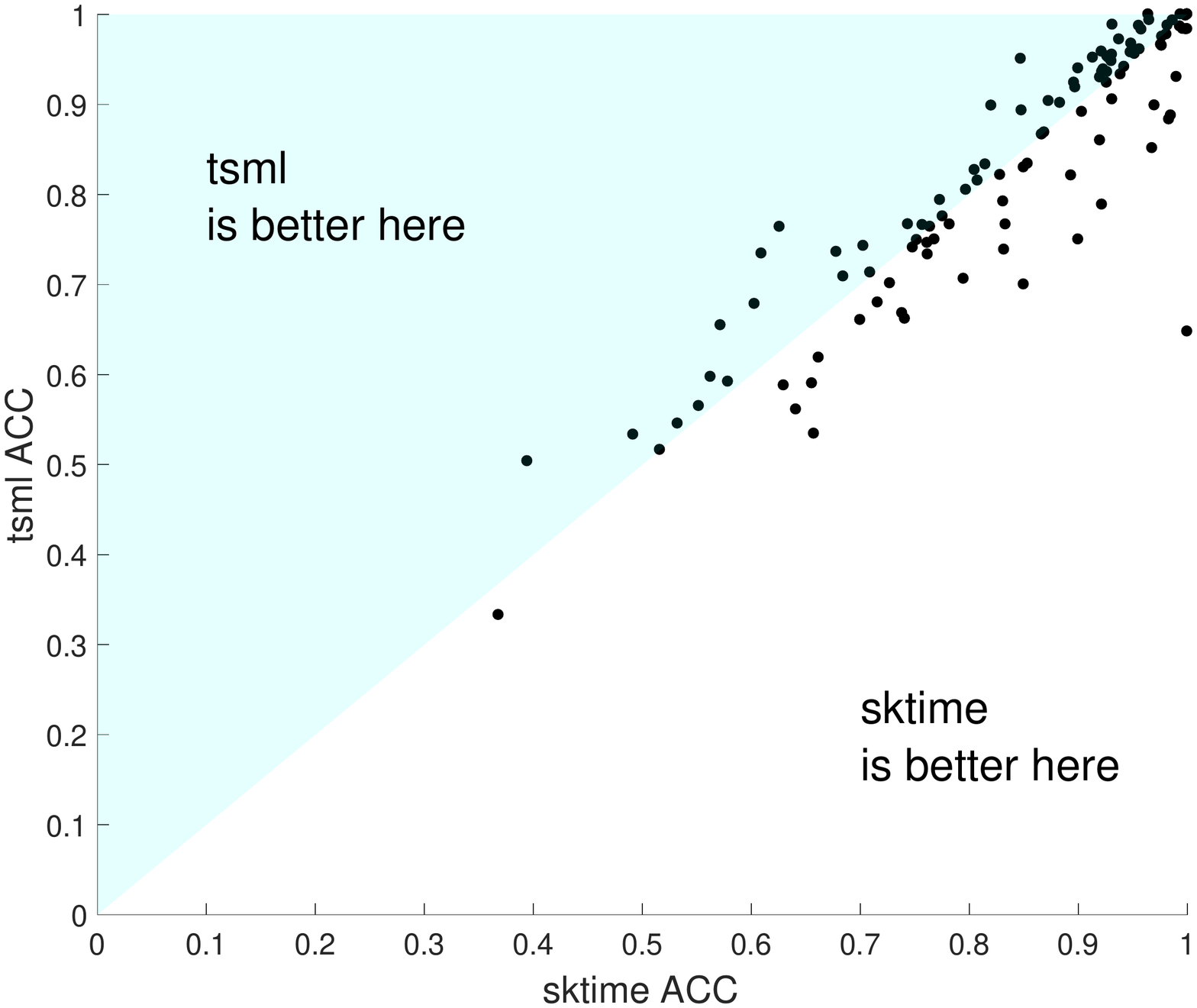}
    \caption{STC sktime vs tsml scatterplot of test accuracies. sktime Wins/Draw/Losses: 48/11/52}
    \label{fig:stc_pairwise}
 \end{figure}

The shapelet transform classifier (\texttt{ShapeletTransformClassifier} in file \texttt{stc.py}) is a single pipeline of a shapelet transform (\texttt{transformers/shapelets.py}) and a classifier (by default, a random forest with 500 trees).  The original shapelet transform performed a complete enumeration of the shapelet space, and used a heterogeneous ensemble for a classifier~\cite{hills14shapelet}. More recent research has shown that equivalent accuracy can be obtained through randomly searching for shapelets rather than a full enumeration~\cite{bostrom17binary}. The shapelet transform in sktime is contractable, in that you can specify an amount of time to search for shapelets before performing the transform. The default is 300 minutes. The latest STC in tsml uses a rotation forest classifier, since it has been shown to be best for this type of problem~\cite{bagnall18rotf}. Rotation forest is not yet available in python (we are working on an alpha version), hence we use random forest for our comparison. We run STC in sktime and tsml with a 1000 minute time limit and use the sklearn amd Weka random forest implementations as the final classifier. Given both classifiers are contracted, we only compare the accuracies. Figure~\ref{fig:stc_pairwise} shows that, overall, there is no significant difference between the two implementations. However, there is still some considerable variation. This may be down to evaluation over a single resample, or related to core differences in the underlying implementation. Further investigation is required.

\section{Distance Based: Elastic Ensemble (EE) and Proximity Forest (PF)}
\label{sec:distance}



\subsection{Elastic Ensemble~\cite{lines15elastic}}

The Elastic Ensemble (EE)~\cite{lines15elastic} ensembles 11 different time series distance measures each coupled with a 1-nearest-neighbour (NN) classifier, together known as the constituents of EE. The variation in distance measure of each constituent injects diversity into the ensemble and provides superior classification performance over any individual constituent alone. EE uses the following distance measures: Euclidean distance, Dynamic Time Warping (DTW), Derivative DTW (DDTW)~\cite{keogh01derivative}, DTW with cross-validated warping window (DTWCV)~\cite{ratanamahatana05threemyths}, DDTW with cross-validated warping window (DDTWCV), Lowest Common Sub-Sequence (LCSS)~\cite{hirschberg77algorithms}, Edit Distance with Real Penalty (ERP)~\cite{chen04erd}, Move-Split-Merge (MSM)~\cite{stefan13move-split-merge}, and Time Warp Edit Distance (TWED)~\cite{marteau09stiffness}. The latter eight of these distance measures benefit from hyper-parameter tuning. By default, EE employs a tuning process for each constituent over 100 parameter options using leave-one-out-cross-validation (LOOCV).

Comparing long time-series using distance measures can be time consuming, therefore both a cython and python version are provided in sktime. These are in \texttt{distances.elastic.py} and \texttt{distances.elastic\_cython.pyx}. Further, sktime provides an extension to scikit-learn's \texttt{KNeighborsClassifier} to  enable usage with pandas dataframes and custom distance measures. The tuning of each constituent of EE can be time consuming also, therefore we provide capability to reduce the number of neighbours used in examination of parameter options and reduce the number of parameters per constituent overall. Both of these have been shown to significantly speed-up EE with no significant loss in classification performance.

\newpage
\begin{Verbatim}[frame=single]
    from sktime.classifiers.distance_based.elastic_ensemble
        import ElasticEnsemble
    from sktime.classifiers.distance_based.time_series_neighbors
        import KNeighborsTimeSeriesClassifier
    from sklearn.model_selection
        import GridSearchCV,
               LeaveOneOut

    # 1NN classifier with full DTW as the distance measure
    dtw_nn = KNeighborsTimeSeriesClassifier(metric='dtw',
        n_neighbors=1)

    # 1NN classifier tuned for best accuracy over 100 DTW
    # warping window options
    dtwcv_nn = GridSearchCV(
        estimator=KNeighborsTimeSeriesClassifier(metric='dtw',
            n_neighbors=1),
        param_grid={'metric_params':
            [{'w': x / 100} for x in range(0, 100)]}
        cv=LeaveOneOut(),
        scoring='accuracy'
    )

    # Default EE with full tuning effort
    ee = ElasticEnsemble()

    # Configuration for reducing tuning efforts of the EE
    ee = ElasticEnsemble(proportion_of_param_options = 0.5,
         proportion_of_train_in_param_finding = 0.1)
\end{Verbatim}
Even with cython, the full Elastic Ensemble is very slow in both absolute and relative terms. In practice, we never run it as a single classifier. Rather, we build each constituent in parallel and ensemble after each is complete.

\begin{figure}[!h]
\centering
    \includegraphics[width=\textwidth, trim={0cm 0cm 0cm  3cm},clip]{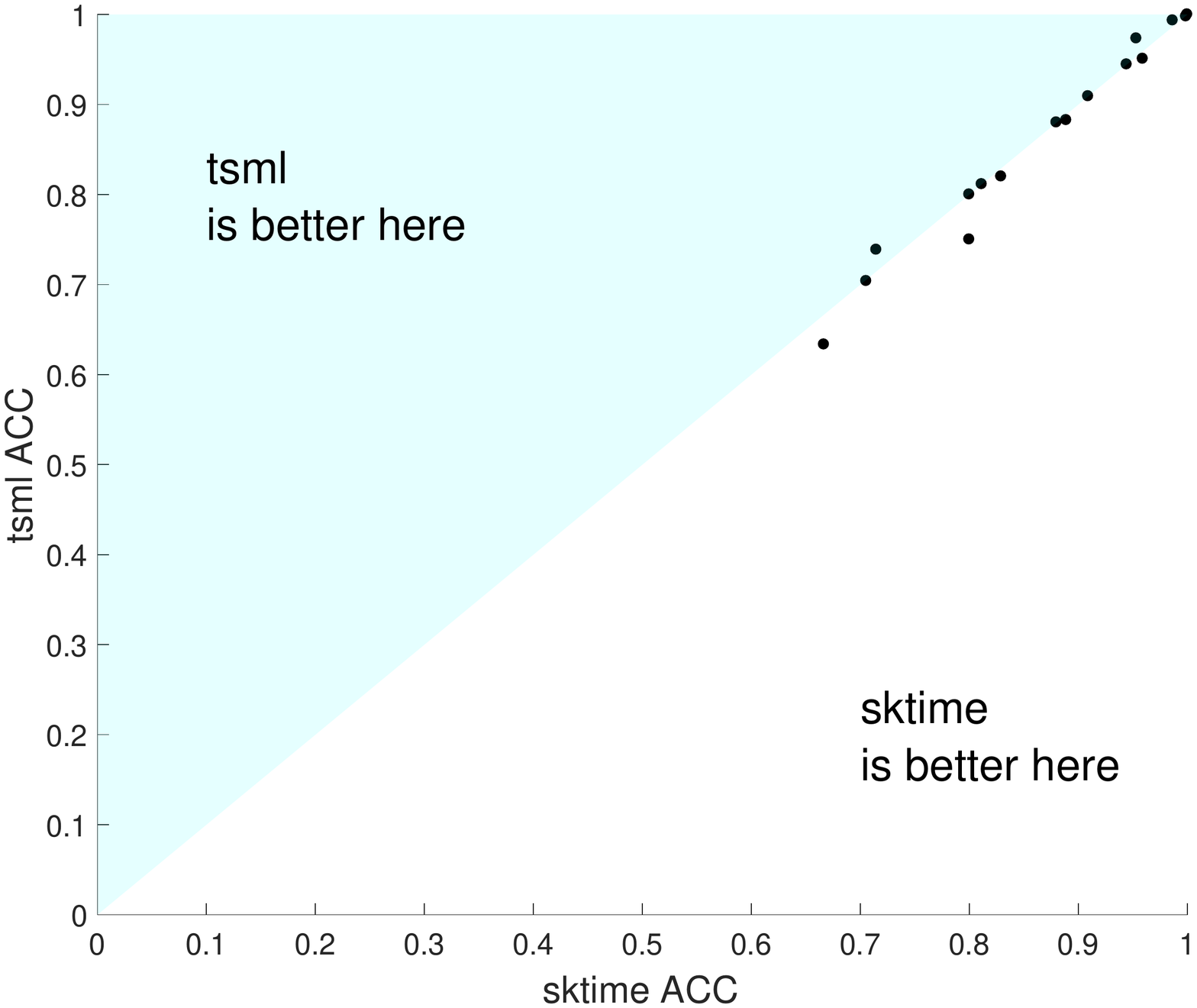}
    \caption{EE sktime vs tsml scatterplot of test accuracies.}
    \label{fig:boss_pairwise}
 \end{figure}
\begin{figure}[!h]
	\centering
        \includegraphics[width=\textwidth, trim={0cm 0cm 0cm  0cm},clip]{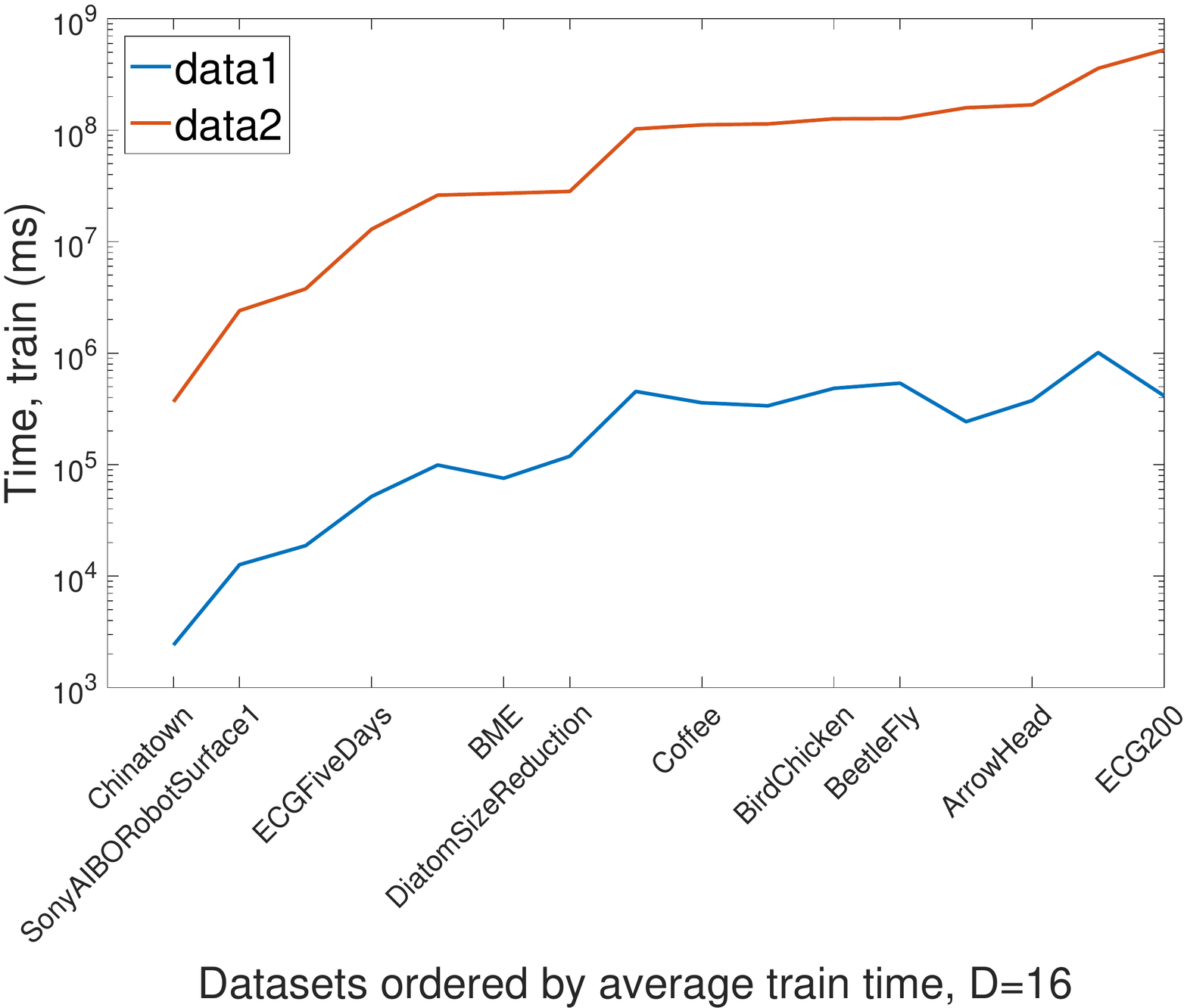}
    \caption{Train time for two EE classifiers: sktime (data2) and tsml (data1).}
    \label{fig:boss_timing}
\end{figure}

\subsection{Proximity Forest}

Proximity Forest (PF)~\cite{lucas19proximity} is a competitor to EE and implements a forest of decision trees, known as Proximity Trees (PT), in a Random Forest like structure. A Proximity Tree (PT) partitions data by comparing the relative similarity of instances against a set of exemplar instances. At any given node in the PT, a one exemplar instance is randomly picked per class from the data passed down from the parent node. The data is partitioned based upon the similarity of data instances against exemplar instances, grouping similar instances. Each group of instances, along with the corresponding exemplar, are passed down as input data to a child node and the process repeats until leaf nodes are pure. This process is known as splitting.

The similarity of instances against exemplar instances is determined using distance measures. PF uses the same distance measures and corresponding parameter ranges as EE. Each split in a PT randomly chooses the distance measure and corresponding parameter set if required. This randomness can lead to poor splits, therefore a hyperparameter $r$ controls the evaluation of multiple splits at each node, choosing the best split based upon gini score. Each split can be seen as a PT of height 1, otherwise known as a Proximity Stump (PS).

The PF generates a set number of PTs and combines their predictions using a majority vote. The extreme randomisation across the PF produces classifier which can be built in a fraction of the time of EE and produce competitive prediction performance.

\begin{Verbatim}[frame=single]
    from sktime.classifiers.distance_based.proximity_forest
        import ProximityForest,
               ProximityStump,
               ProximityTree

    # Proximity Stump (PT with 1 level of depth)
    ps = ProximityStump()

    # Proximity Tree with 5 split evaluations per tree node
    pt = ProximityTree(n_stump_evaluations = 5)

    # Proximity Forest with 100 trees and
    #   5 split evaluations per tree node
    pf = ProximityForest(n_trees = 100, n_stump_evaluations = 5)
\end{Verbatim}

\begin{figure}[!h]
\centering
    \includegraphics[width=\textwidth, trim={0cm 0cm 0cm  3cm},clip]{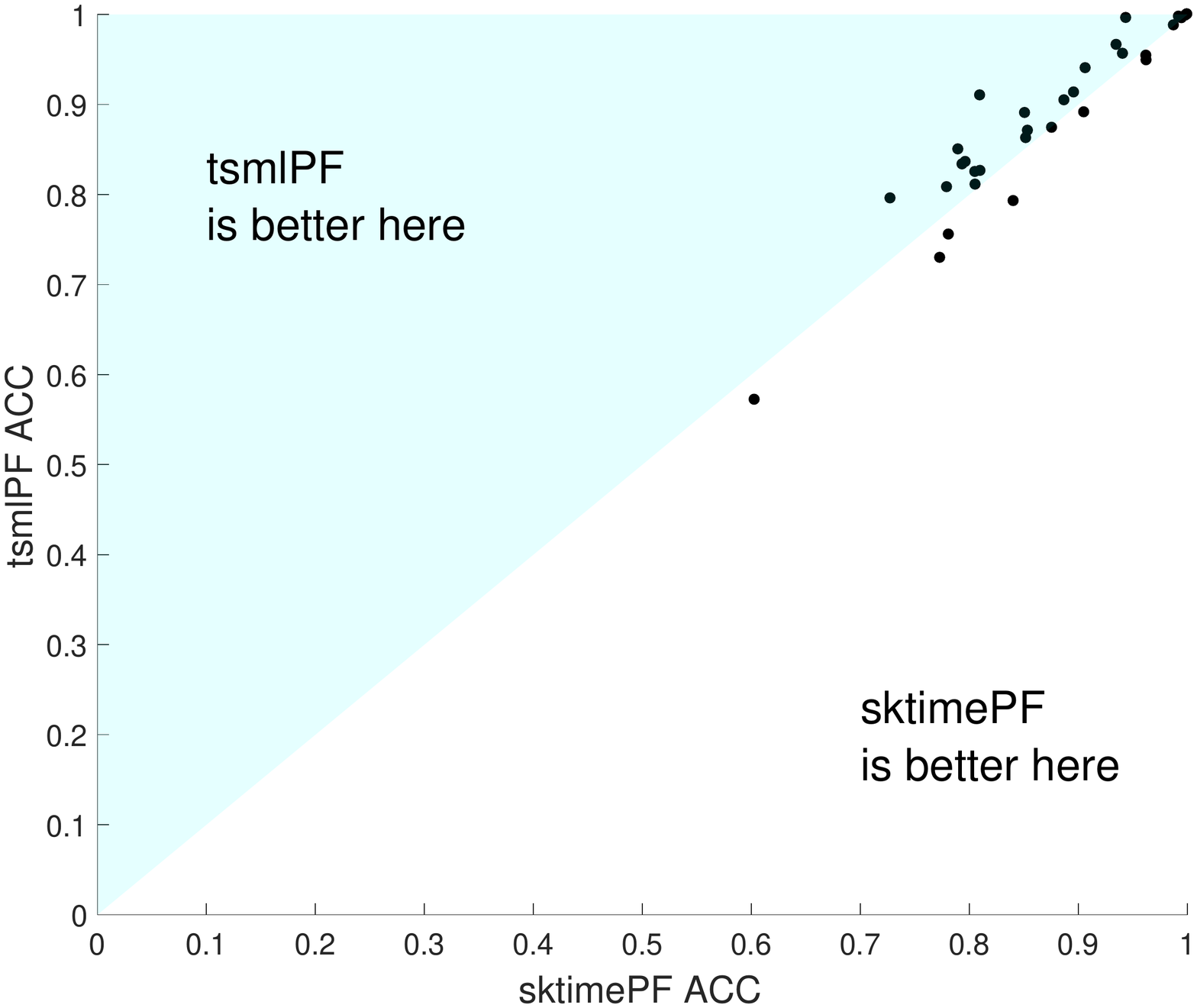}
    \caption{PF sktime vs tsml scatterplot of test accuracies.}
    \label{fig:boss_pairwise}
 \end{figure}
\begin{figure}[!h]
	\centering
        \includegraphics[width=\textwidth, trim={0cm 0cm 0cm  0cm},clip]{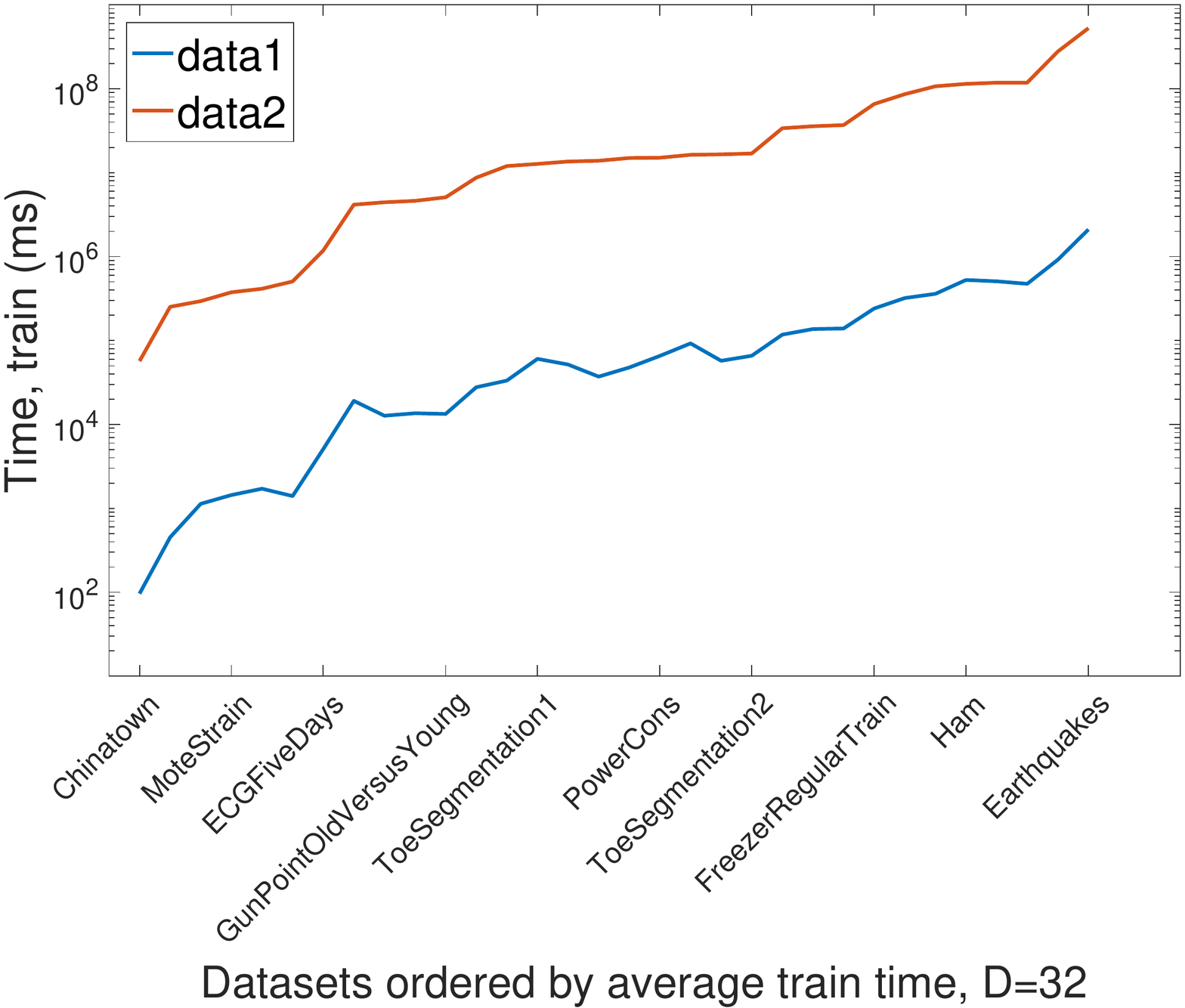}
    \caption{Train time for two PF classifiers: sktime (data2) and tsml (data1).}
    \label{fig:boss_timing}
\end{figure}

\section{Conclusions and Future Directions}
Developing time series classification algorithms is an active research field. Our goal in providing sktime is to make researching this field easier and more transparent. If researchers use a common framework, reproducability and comparison to the current state of the art becomes easy. In this way we hope to drive the field forward to help automatically answer the key questions for classifying time series: which is the best algorithm for the task at hand.

The classification component of sktime has broad functionality, but there is much that could be added and improved. Our work plan for the medium term is as follows.

\subsection{Improved Classifier Functionality}
It has taken a surprising amount of effort to get this far, but there is still much to do.
This exercise has highlighted several issues with sktime that need resolving before the next release to master.
\begin{enumerate}
    \item interval based: why is the composite version so much slower than the others?
    \item frequency based: can we speed up RISE through matrix operations without exception being continually thrown?
    \item dictionary based: can we speed up BOSS through a more pythonesque implementation?
    \item shapelet based: can we improve accuracy by using rotation forest and improving the search efficiency through cython?
    \item Can we improve EE performance through restricting the parameter search?
\end{enumerate}
We also intend to improve the functionality by making all classifiers contractable and allow for efficient estimation of the test error from the train data. In addition to improving the existing set of classifiers in these ways, we intend porting in recently proposed alternative TSC algorithms such as FastEE, TS-CHIEF~\cite{shifaz19ts-chief}, WEASEL~\cite{schafer2017fast}, Shapelet Forest~\cite{karlsson16generalized} and HIVE-COTE~\cite{lines18hive}. We have implemented the range of distance functions and kernels described in~\cite{abanda19distance} and the Matrix Profile distance MPDist~\cite{gharghabi18matrixprofile12}. We will evaluate these classifiers to examine how they perform in relation to those already implemented within the toolkit.

\subsection{Handle Unequal Length Series}
One of the primary motivators for our data model of pandas of series objects was to facilitate the easy incorporation of unequal length time series as classification problems. The current suite of available algorithms needs to be adjusted to allow for this use case. For some algorithms this will be fairly simple: BOSS, for example, can simply normalise the histograms and shapelets are independent of series length. However, these design choices may have an impact on accuracy and experimentation using the enhancements will be required.

\subsection{Multivariate Time Series Classifiers}
sktime already provides two composition interfaces for solving multivariate time series classification problems, including
\begin{itemize}
    \item the \texttt{ColumnConcatenator}, a transformer that can be used to concatenate two or more time series columns into a single long time series column, so that one can then apply a classifier to the concatenated, univariate data, and
    \item the \texttt{ColumnEnsembleClassifier}, a meta-estimator that allows for column-wise ensembling in which a separate classifier is fitted for each time series column and their predictions are aggregated.
\end{itemize}
We have started to implement bespoke methods for specific estimators to handle multivariate time series input data and we aim to add more bespoke methods in future work, e.g.\ finding shapelets in multidimensional space or extending time series forest to extract features from multidimensional segments (or hyper-planes). There are also a multitude of bespoke methods for classifying time series. We shall draw up a candidate

\subsection{Time series regression}
The composite structure of many time series classifiers allows us to refactor them into their regressor counterparts. We have started implementing regression algorithms, including a TSF regressor and are currently working on adding more regressors.

\section*{Acknowledgements}{
 We would like to thank sktime contributors who are not specific authors on this paper and our colleagues at UCR with whom we maintain the time series data repositories.

 \bibliographystyle{plain}

\end{document}